# Financial Text Classification Based On rLoRA Finetuning On Qwen3-8B model

Zhiming Lian

LL Funds LLC, Philadelphia, United States, zhiming.lian@llfunds.com

**Abstract**

Financial text classification has increasingly become an important aspect in quantitative trading systems and related tasks, such as financial sentiment analysis and the classification of financial news. In this paper, we assess the performance of the large language model Qwen3-8B on both tasks. Qwen3-8B is a state-of-the-art model that exhibits strong instruction-following and multilingual capabilities, and is distinct from standard models, primarily because it is specifically optimized for efficient fine-tuning and high performance on reasoning-based benchmarks, making it suitable for financial applications. To adapt this model, we apply Noisy Embedding Instruction Finetuning and based on our previous work, this method increases robustness by injecting controlled noise into the embedding layers during supervised adaptation. We improve efficiency further with Rank-stabilized Low-Rank Adaptation low-rank optimization approach, and FlashAttention, which allow for faster training with lower GPU memory. For both tasks, we benchmark Qwen3-8B against standard classical transformer models, such as T5, BERT, and RoBERTa, and large models at scale, such as LLaMA1-7B, LLaMA2-7B, and Baichuan2-7B. The findings reveal that Qwen3-8B consistently surpasses these baselines by obtaining better classification accuracy and needing fewer training epochs. The synergy of instruction-based fine-tuning and memory-efficient optimization methods suggests Qwen3-8B can potentially serve as a scalable, economical option for real-time financial NLP applications. Qwen3-8B provides a very promising base for advancing dynamic quantitative trading systems in the future.

**CCS Concepts**

Computing methodologies -> Artificial intelligence -> Natural language processing -> Natural language generation

**Additional Keywords and Phrases**

Financial Text Classification, rLoRA, Qwen3-8B, LLaMA

## 1 INTRODUCTION

Text classification is an essential task in natural language processing (NLP), which aims to categorize a piece of text, such as a sentence, document or message, into a pre-defined category. This task is a foundational approach to many subsequent applications that include sentiment analysis, topic classification, spam detection, intent classification, and news classification. Traditional text classification relied primarily on statistical or linguistic approaches. For instance, the bag-of-words and n-gram models, while simple, capture the word frequency and contextual patterns of text representation. The term frequency-inverse document frequency (TF-IDF) introduced extended the bag-of-words or n-gram representation by down-weighting common words and focusing on identifying more discriminating features [1]. These models are computationally cheap and easy to interpret which is helpful for baseline models or modelling in resource-constrained situations. The availability of distributed representations in language has greatly improved the field of text classification. For example, word embeddings (e.g., Word2Vec, GloVe) provide representations of words in denser spaces where models can observe semantic similarity and syntactic regularities in word use [2,3]. This increased the ability of classifiers (e.g., logistic regression, support vector machine) to use more meaningful features instead of the sparse vector representation. Also, recurrent neural networks (RNNs), especially long short-term memory (LSTM) or gated recurrent units (GRUs) networks, added additional advances to the field of discovery through modeling sequential dependencies and long-range context, which is a vital component to tasks such as document level sentiment classification or opinion mining [4,5].

Although these enhancements have improved text classification, recurrent models often continue to struggle with vanishing gradients and ineffective parallelization. Hybrid approaches combined embeddings with convolutional neural networks (CNNs) and built local feature reduction, while hierarchical architecture processed text at various levels (i.e. word, sentence, paragraph). This led to more complicated and powerful models. Although transformers and large-scale pre-trained models dominate text classification research today, traditional methods are still relevant. In many real-world contexts (i.e. low-resource contexts, imbalanced datasets, or applications where model interpretation is needed), methods such as n-grams, TF-IDF, and LSTM-based classifiers can be effective and inexpensive.

## 2  RELATED WORK

Text classification has undergone a transformation from naive frequency-based representations to more developed large-scale pre-trained models. Early deep learning models were based on recurrent architectures such as LSTM and GRU, which were effective in modeling sequential dependencies with text. Subsequently, transformer architecture fundamentally changed the course of natural language processing (NLP) by utilizing parallel computation and more effectively modeling long-range dependencies [6].

BERT (Bidirectional Encoder Representations from Transformers) was one of the first transformer-based large-scale models to show large performance improvements across a variety of NLP benchmarks [7]. RoBERTa built upon BERT by removing next sentence prediction and introducing dynamic masking with larger datasets, which improved performance in sentiment analysis, question answering, and classification [8]. T5 (Text-to-Text Transfer Transformer) unified a framework with a modified premise where all NLP tasks are reformulated as text-to-text (questions are asked as text for segmentation) tasks, showing state-of-the-art or better superior performance on multiple benchmarks through classification and summarization [9].

In addition to those foundational architectures, open-source large language models have increased the potential of tasks around text classification. LLaMA (Large Language Model Meta AI) was a resource-efficient option with competitive multi-lingual capabilities [10] that allowed practitioners without in-depth machine learning expertise to perform tasks in many languages. LLaMA2 expanded on this success by developing LLaMA's architecture, better training approaches, and increased token counts, all of which resulted in superior performance on a variety of classification and reasoning benchmarks [11]. Baichuan2 is another open-source model that arose to compete with LLaMA2, which was trained on trillions of tokens and was particularly useful for multilingual tasks, such as sentiment and topical classification [12].

These models illustrate the shift from using smaller, domain-adapted models to general-purpose large language models, which provide general capabilities for addressing a diverse engagement of different text classification tasks. While traditional architecture, such as BERT or RoBERTa, represents strong baselines for performance, the use of large-scale architecture represented by LLaMA and Baichuan2 will change the landscape of how we perform text classification in regards to improved accuracy and scalability.

## 3  METHODOLOGY

### 3.1 Qwen3 and Qwen3-8B architecture & innovations

We opt for the Qwen3 family as our baseline due to its consolidated "thinking / non-thinking" inference modes and robust multilingual pretraining. Specifically, Qwen3-8B is a dense causal decoder equipped with grouped-query attention (GQA), rotary position embeddings (RoPE), and resilient multi-linearity factorization. The official technical report outlines the dual-mode design (with thinking tokens budgeted) and native 32K context window scalable beyond, the official report discusses the dual-mode design (where thinking tokens are budgeted) and native 32,768 context window was forgivingly scalable beyond, which is helpful to parse long-form financial texts [14]. Qwen3-8B has 8.2 billion parameters and 36 transformer layers. Grouped query attention (GQA) uses 32 query heads and 8 key-value heads. The hidden size is made consistent to optimize KV-cache, reducing latency in inference with negligible loss of accuracy. Wthe native context window is 32,768 tokens, but extends to 131,072 tokens in evaluation by scaling techniques such as YaRN. These decisions promote efficiency in reasoning while retaining accuracy across challenging NLP benchmarks. In practice, we turn on thinking mode for chain-of-thought heavy subtasks (e.g., ambiguous sentiment involving numerical data), and turn it off for classification tasks performed in batch, per the Qwen3 documentation. The architecture can be seen in Figure 1 and the concrete parameters are listed in Table 1.

Table 1. Architecture parameters of Qwen3-8B

| Hyperparameter | Qwen3-8B Value |
| --- | --- |
| Model size (Parameters) | ~8.2B |
| Transformer layers | 36 |
| Hidden size | 5,120 |
| Intermediate size (FFN) | 13,696 |
| Attention heads | 32 |
| Key/Value heads (GQA) | 8 |
| Attention type | Grouped-Query Attention (GQA) |
| Normalization | RMSNorm |
| Positional encoding | Rotary Position Embedding (RoPE) |
| Vocabulary size | 151,552 |
| Context length | 32,768 (native), up to 131,072 |
| Activation function | SwiGLU |
| Dropout | 0 (inference) |
| Precision | bfloat16 / float16 (mixed precision) |

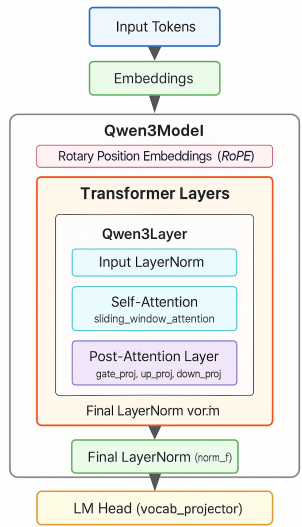

Figure 1. Architecture of Qwen3

### 3.2 rLoRA technique for finetuning

For our parameter-efficient finetuning, we employ LoRA (shown in Figure 2) and its rank-stabilized version. LoRA adds trainable low-rank matrices into the weight projections we selected, while all other model weights remain frozen. This approach uses less GPU memory, allowing us to finetune very large models with lower compute resources [15].

The original LoRA is usually unstable when the rank parameter is increased. This problem is effectively resolved through rank-stabilized LoRA (rLoRA), whereby the updates are scaled with $\sqrt{r}$, so that both stability and convergence are improved. This means we can use higher ranks safely without hurting performance when possible, especially for classifiers doing difficult classification subtasks. rLoRA uniquely beats baseline LoRA across the board while remaining efficient enough to use in a scale for finetuning in the financial domain [16].

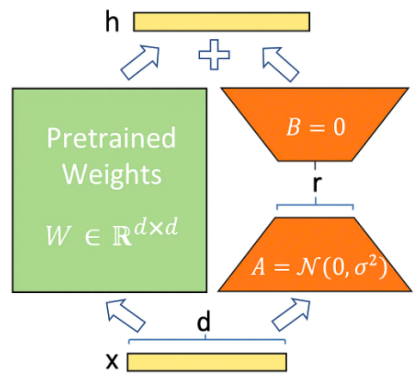

Figure 2. LoRA technique

### 3.3 Instruction finetuning and template

Consistent with the typical instruction-tuning paradigm, we fine-tune Qwen3-8B with carefully-prepared instruction – response pairs. Each instance is a two-turn interaction: the first turn consists of an explicit instruction and the financial text, while the second turn contains the expected categorical label. We have used a template that explicitly controls Qwen3's thinking mode with the \no_think tag. In this way, we can attenuate the input and output token number. This ensures deterministic classification with less latency, rather than verbose reasoning chains. For sentiment classification, we used the following template in Table 2:

Table 2. Financial Text Sentiment Classification Template

| Rows | Detail |
| --- | --- |
| Input Role | user |
| Input Content | \no_think Do sentiment classification for financial text. {Input Text} |
| Output Role | Assistant |
| Output Content | {Output Text} // Netural or Positive or Negative |

## 4  EXPERIMENTS

### 4.1 Financial Dataset

In this research, we utilize datasets by conducting experiments and analyses with two datasets (Seen in Figure 3 & Figure 4). One is a financial sentiment classification dataset with 2,879 neutral, 1,362 positive, and 604 negative samples, resulting in a reasonably balanced distribution of sentiment categories. The second dataset is a Twitter-based financial news dataset publicly available on Kaggle that we use for model fine-tuning to classify financial news articles into 20 categories. The Twitter dataset is split into a training dataset of 16,990 samples and a test dataset of 4,117 samples. When considered together, the financial news datasets represent sufficiently high scale and diversity to use for instruction fine-tuning to adjust the performance of large language models to financial text classification.

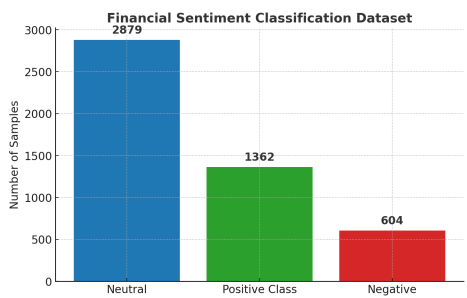

Figure 3. Financial Sentiment Text Distribution

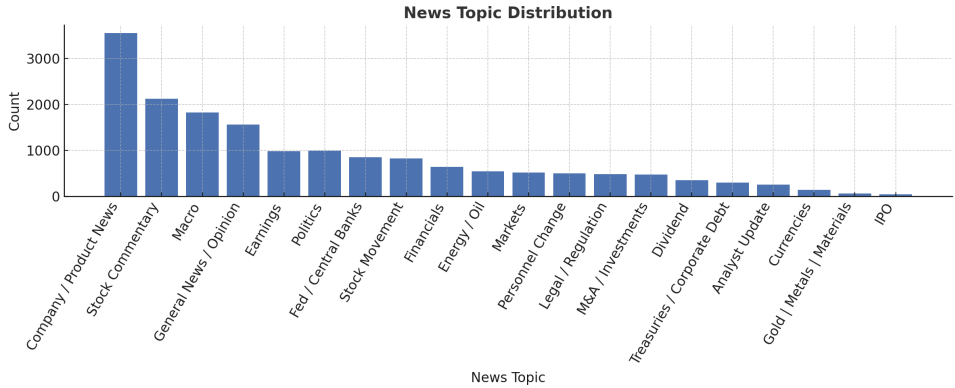

Figure 4. Financial Topic Text Distribution

### 4.2 Experiment Settings

In our experiments, we are using Qwen3-8B as the backbone model because of its grouped-query attention and long-context capabilities. The finetuning process was conducted in a supervised instruction-tuning fashion, where the model was finetuned to match target labels with user instructions for financial sentiment and news classification tasks. For increased robustness and generalizability, we implemented Noisy Embedding Instruction Finetuning (NEFTune), which enables the addition of small perturbations to embedding representations during training. NEFTune uses stochastic noise to regularize the embedding space while maintaining the embedding representations of user instructions and labels. This method reduces overfitting and improves instruction following ability, while maintaining the same computational budget. In our experiment, the NEFTune noise scale parameter ($\alpha$) was set to 0.3, which adequately balances the amount of stable with diversity.

Table 3: Experiment Settings

| Data | Description |
| --- | --- |
| Batch Size | 3 |
| Gradient Accumulation | 4 |
| Learning Rate | 5e-5 |
| Optimizer | Adam |
| Max token length | 360 |
| Use rLoRA | True |
| Lora Rank | 8 |
| Lora Dropout | 0.1 |
| Epoch | 3 |
| NEFTune alpha | 0.3 |
| Use FlashAttention | True |

Table 3 provides a brief summary of the training hyperparameters. In particular, we use a batch size of 3 with a gradient accumulation of 4 steps. This leads to an effective batch size of 12. We fix the learning rate to 5e-5 and optimize it with Adam. The maximum token length is capped at 360 in order to accommodate relatively long financial texts. For parameter-efficient tuning, we also enable rLoRA with a rank of 8 and a dropout of 0.1. The rLoRA will stabilize the scaling of low-rank adapters so that the updates might be effective even for a larger rank. We train on 3 epochs, which is sufficient to converge through the dataset sizes.

We also incorporate FlashAttention, which speeds up the computation of all aspects of attention with less memory usage that enables us to keep a high throughput given that the sequence length is relatively long. Overall, these provide an overall balanced configuration in efficiency and stability for large scale financial instruction finetuning.

### 4.3 Experiments result

Table 4. Evaluation of different methods on accuracy metric

| Models | Model Type | Financial Sentiment Classification (ACC) | Financial Toipc Classification (ACC) |
| --- | --- | --- | --- |
| Roberta | Not LLM | 0.7928 | 0.8612 |
| Bert | Not LLM | 0.7854 | 0.8523 |
| Baichuan2-7B | LLM | 0.8165 | 0.8784 |
| Llama-7B | LLM | 0.8297 | 0.8801 |
| Llama2-7B | LLM | 0.8322 | 0.8877 |
| **Qwen3-8B** | **LLM** | **0.8415** | **0.9315** |

Table 4 provides accuracy metrics for both financial sentiment classification and financial topic classification. Non-LLM baseline classification approaches performed quite well, but somewhat limited; RoBERTa received an accuracy of 0.7928 and 0.8612 for sentiment and topics, respectively, while BERT received an accuracy of 0.7854 and 0.8523. This indicates that pre-trained transformers can learn about the domain but struggle with the complexity of instruction-following tasks. LLM-based approaches showed clear advantages; Baichuan2-7B achieved accuracy scores of 0.8165 for sentiment classification and 0.8784 for topic classification. The results of LLaMA-7B and LLaMA2-7B were similar, receiving a sentiment score of 0.8297/0.8801 and 0.8322/0.8877, respectively. The consistent positive results we see here help to reinforce that scaling model ability and pretraining strategies are useful for domain adapted accurate classification tasks. Qwen3-8B achieves better performance than all baselines, reaching 0.8415 on sentiment classification and 0.9315 on topic classification.

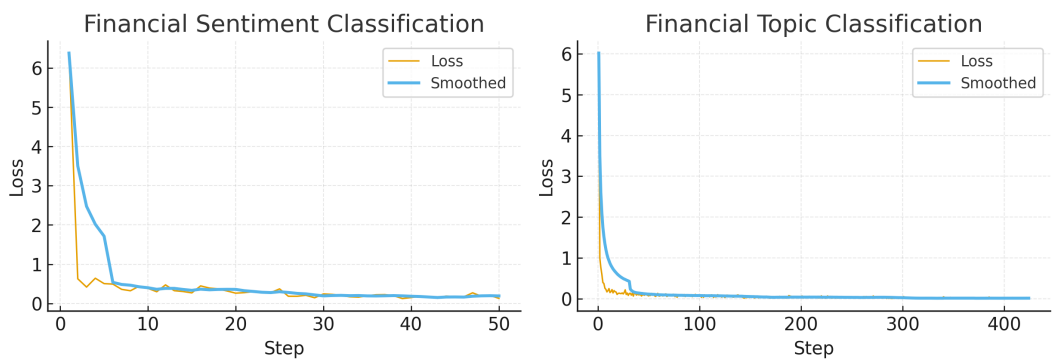

Figure 5. Training Loss Iteration Plot

In Figure 5, our training approach demonstrates rapid convergence, achieving stable performance within only three epochs and requiring relatively few iterations. In contrast, traditional non-LLM methods typically need more than ten epochs to reach comparable results.

## 5 CONCLUSION

In this study, we explored how Qwen3-8B could perform in the context of financial text classification as it relates to sentiment and topic classification. In doing so, we proposed a training approach that utilizes Noisy Embedding Instruction Finetuning in conjunction with both rLoRA and FlashAttention, which produces an effecient and robust finetuning strategy. We find that Qwen3-8B quickly converges with stable performance after three epochs of fine-tuning whilst requiring relatively fewer iterations, while prior non-LLM approaches usually take over ten epochs of tuning to achieve a similar accuracy for financial classification. Furthermore, our evaluation results indicate that Qwen3-8B outperformed both classical transformer baselines as well other large language models (LLMs) across both tasks with higher task accuracies. This performance improvement was a product of instruction-based finetuning, noise-regularized embeddings and parameter-efficient finetuning performed in conjunction with each other to improve generalization whilst lowering the computation cost for both tasks. Overall, these findings suggest that Qwen3-8B not only work on financial NLP benchmarks but it is also practical and scaleable to real-world applications such as quantitative trading systems where accuracy, efficiency, and adaptability are paramount.